\documentclass[onecolumn,11pt]{article}

\usepackage[sc,osf]{mathpazo} 
\usepackage{eulervm}

\usepackage{simplemargins}
\usepackage{graphicx}
\usepackage{enumitem}
\usepackage{url}
\usepackage{amsmath}
\usepackage{amssymb}
\usepackage{cite}

\setleftmargin{2cm}
\setrightmargin{2cm}
\settopmargin{2cm}
\setbottommargin{2cm}

\newcommand{\Mod}[1]{\ (\mathrm{mod}\ #1)}
\newcommand{\igjskill}{\textsc{skill}}
\newcommand{\igjstamina}{\textsc{stamina}}
\newcommand{\igjluck}{\textsc{luck}}

\title{\textbf{Optimal strategies in the Fighting Fantasy gaming system: influencing stochastic dynamics by gambling with limited resource} \\ \vspace{0.5cm} \normalsize Iain G. Johnston${}^{1,2}$ \\ \vspace{0.5cm}  \normalsize ${}^1$ Department of Mathematics, Faculty of Mathematics and Natural Sciences, University of Bergen, Bergen, Norway \\ \normalsize ${}^2$ Alan Turing Institute, London, UK}
\date{}

\begin{document}
\maketitle

\section*{Abstract}
Fighting Fantasy is a popular recreational fantasy gaming system worldwide. Combat in this system progresses through a stochastic game involving a series of rounds, each of which may be won or lost. Each round, a limited resource (`\igjluck{}') may be spent on a gamble to amplify the benefit from a win or mitigate the deficit from a loss. However, the success of this gamble depends on the amount of remaining resource, and if the gamble is unsuccessful, benefits are reduced and deficits increased. Players thus dynamically choose to expend resource to attempt to influence the stochastic dynamics of the game, with diminishing probability of positive return. The identification of the optimal strategy for victory is a Markov decision problem that has not yet been solved. Here, we combine stochastic analysis and simulation with dynamic programming to characterise the dynamical behaviour of the system in the absence and presence of gambling policy. We derive a simple expression for the victory probability without \igjluck{}-based strategy. We use a backward induction approach to solve the Bellman equation for the system and identify the optimal strategy for any given state during the game. The optimal control strategies can dramatically enhance success probabilities, but take detailed forms; we use stochastic simulation to approximate these optimal strategies with simple heuristics that can be practically employed. Our findings provide a roadmap to improving success in the games that millions of people play worldwide, and inform a class of resource allocation problems with diminishing returns in stochastic games.
\\
\textbf{Keywords:} stochastic game; Markov decision problem; stochastic simulation; dynamic programming; resource allocation; stochastic optimal control; Bellman equation


\section*{Introduction}
Fantasy adventure gaming is a popular recreational activity around the world. In addition to perhaps the best-known `Dungeons and Dragons' system \cite{gygax1974dungeons}, a wide range of adventure gamebooks exist, where a single player takes part in an interactive fiction story \cite{costikyan2007games}. Here, the reader makes choices that influence the progress through the book, the encounters that occur, and the outcomes of combats. In many cases, die rolls are used to provide stochastic influence over the outcomes of events in these games, particularly combat dynamics. These combat dynamics affect the game outcome, and thus the experience of millions of players worldwide, yet have rarely been studied in detail. 

Here we focus on the stochastic dynamics of a particular, highly popular franchise of gamebooks, the Fighting Fantasy (FF) series \cite{green2014you}. This series, spanning over 50 adventure gamebooks (exemplified by the famous first book in the series, `The Warlock of Firetop Mountain' \cite{jackson2002warlock}), has sold over $20$ million copies and given rise to a range of associated computer games, board games, and apps, and is currently experiencing a dramatic resurgence in popularity \cite{bbc, osterberg2008rise}. Outside of recreation, these adventures are used as pedagogical tools in the huge industry of game design \cite{zagal2015fighting} and in teaching English as a foreign language \cite{philips1994role}. 

In FF, a player is assigned statistics (\igjskill{} and \igjstamina), dictating combat proficiency and endurance respectively. Opponents are also characterised by these combat statistics. Combat proceeds iteratively through a series of `attack rounds'. In a given round, according to die rolls, the player may draw, win or lose, respectively. These outcomes respectively have no effect, damage the opponent, and damage the player. The player then has the option of using a limited resource (\igjluck{}) to apply control to the outcome of the round. This decision can be made dynamically, allowing the player to choose a policy based on the current state of the system. However, each use of \igjluck{} is a gamble \cite{maitra2012discrete, dubins1965inequalities}, where the probability of success depends on the current level of the resource. If this gamble is successful, the player experiences a positive outcome (damage to the opponent is amplified; damage to the player is weakened). If the gamble is unsuccessful, the player experiences a negative outcome (damage to the opponent is weakened, damage to the player is amplified). The optimal strategy for applying this control in a given state has yet to be found. 

This is a stochastic game \cite{shapley1953stochastic, adlakha2015equilibria} played by one or two players (the opponent usually has no agency to decide strategies) on a discrete state space with a finite horizon. The game is Markovian: in the absence of special rules, the statistics of the player and opponent uniquely determine a system state, and this state combined with a choice of policy uniquely determine the transition probabilities to the next state. The problem of determining the optimal strategy is then a Markov decision problem (MDP) \cite{bellman1957markovian, kallenberg2003finite}. In an MDP, a decision-maker must choose a particular strategy for any given state of a system, which evolves according to Markovian dynamics. In FF combat, the decision is always binary: given a state, whether or not to use the dimishing resource of \igjluck{} to attempt to influence the outcome of a given round. 

The study of stochastic games and puzzles is long established in operational research \cite{smith2007dynamic, bellman1965application} and has led to several valuable and transferrable insights \cite{smith2007dynamic, little1963algorithm}. Markov analysis, dynamic programming, and simulation have been recently used to explore strategies and outcomes in a variety of games, sports, and TV challenges \cite{lee2012comparison, smith2007dynamic, johnston2016endless, perea2007dynamic, percy2015strategy, clarke2003dynamic}. Specific analyses of popular one-player recreational games with a stochastic element including Solitaire \cite{rabb1988probabilistic, kuykendall1999analyzing}, Flip \cite{trick2001building}, Farmer Klaus and the Mouse \cite{campbell2002farmer}, Tetris \cite{kostreva2003multiple}, and The Weakest Link \cite{thomas2003best}. These approaches typically aim to identify the optimal strategy for a given state, and, in win/lose games, the overall probability of victory over all possible instances of the game \cite{smith2007dynamic}. In stochastic dynamic games, counterintuitive optimal strategies can be revealed through mathematical analysis, not least because `risking points is not the same as risking the probability of winning' \cite{neller2004optimal}. 

The FF system has some conceptual similarities with the well-studied recreational game Pig, and other so-called `jeopardy race games' \cite{neller2004optimal, smith2007dynamic}, where die rolls are used to build a score then a decision is made, based on the current state of the system, whether to gamble further or not. Neller \& Presser have used a value iteration approach to identify optimal strategies in Pig and surveyed other similar games \cite{neller2004optimal}. In FF combat, however, the player has potential agency both over their effect on the opponent and the opponent's effect on them. Further, resource allocation in FF is a dynamic choice and also a gamble \cite{maitra2012discrete, dubins1965inequalities}, the success probability of which diminishes as more resource is allocated. The probability of a negative outcome, as opposed to a positive one, therefore increases as more resource is used, providing an important `diminishing returns' consideration in policy decision \cite{deckro2003modeling}. In an applied context this could correspond to engaging in, for example, espionage and counterespionage \cite{solan2004games}, with increasing probability of negative outcomes with more engagement in these covert activites.

The optimal policy for allocating resource to improve a final success probability has been well studied in the context of research and development (R\&D) management \cite{heidenberger1999research, baye2003strategic, canbolat2012stochastic, gerchak1999allocating}. While policies in this field are often described as `static', where an initial `up-front' decision is made and not updated over time, dynamic policy choices allowing updated decisions to be made based on the state of the system (including the progress of competitors) have also been examined \cite{blanning1981variable, hopp1987sequential, posner1990optimal}. Rent-seeking `contest' models \cite{clark1998contest} also describe properties of the victory probability as a function of an initial outlay from players. The `winner takes all' R\&D model of Canbolat \emph{et al.}, where the first player to complete development receives all the available payoff, and players allocate resource towards this goal \cite{canbolat2012stochastic}, bears some similarity to the outcomes of the FF system. The model of Canbolat \emph{et al.} did not allow dynamic allocation based on the current system state, but did allow a fixed cost to be spread over a time horizon, and computed Nash equilibria in a variety of cases under this model. 

A connected branch of the literature considers how to allocate scarce resource to achieve an optimal defensive outcome \cite{golany2015allocating, valenzuela2015multiresolution}, a pertinent question both for human \cite{golany2009nature} and animal \cite{clark1992inducible} societies. Both optimisation and Nash equilibrium approaches are used in these contexts to identify solutions to the resource allocation problem under different structures \cite{golany2015allocating, valenzuela2015multiresolution}. The FF system has such a defensive component, but the same resource can also be employed offensively, and as above takes the unusual form of a gamble with a diminishing success probability.

We will follow the philosophy of these optimisation approaches to identify the optimal strategy for allocating resource to maximise victory probability from a given state in FF. We first describe the system and provide solutions for victory probability in the case of no gambling and gambling according to a simple policy. Next, to account for the influence of different gambling policies on the stochastic dynamics of the game, we use a backwards-induction approach to solve the Bellman equation for the system and optimise victory probability from any given state. We then identify simple heuristic rules that approximate these optimised strategies and can be used in practise. 

\subsection*{Game dynamics}
Within an FF game, the player has nonnegative integer statistics called \igjskill{}, \igjstamina, and \igjluck{}. \igjskill{} and \igjluck{} are typically $\leq 12$; \igjstamina{} is typically $\leq 24$, although these bounds are not required by our analysis. In a given combat, the opponent will also have \igjskill{} and \igjstamina{} statistics. We label the \igjskill{}, \igjstamina, and \igjluck{} of the player (the `hero') as $k_h, s_h,$ and $l$ respectively, and the opponent's \igjskill{} and \igjstamina{} as $k_o$ and $s_o$. Broadly, combat in the FF system involves a series of rounds, where differences in \igjskill{} between combatants influences how much \igjstamina{} is lost in each round; when one combatant's \igjstamina{} reaches zero or below, the combat is over and that combatant has lost. The player may choose to use \igjluck{} in any given round to influence the outcome of that round. More specifically, combat proceeds through Algorithm 1. \vspace{0.2cm}

\textbf{Algorithm 1. FF combat system.}
\begin{enumerate}[nosep]
\item Roll two dice and add $k_h$; this is the player's attack strength $A_h$.
\item Roll two dice and add $k_o$; this is the opponent's attack strength $A_o$.
\item If $A_h = A_o$, this attack round is a draw. Go to 6.
\item If $A_h > A_o$, the player has won this attack round. \emph{Make decision} whether to use \igjluck{}.
  \begin{enumerate}[noitemsep]
    \item If \emph{yes}, roll two dice to obtain $r$. If $r \leq l$, set $s_o = s_o - 4$. If $r > l$, set $s_o = s_o - 1$. For either outcome, set $l = l - 1$. Go to 6.
    \item If \emph{no}, set $s_o = s_o - 2$. Go to 6.
  \end{enumerate}
\item If $A_h < A_o$, the opponent has won this attack round. \emph{Make decision} whether to use \igjluck{}.
  \begin{enumerate}[noitemsep]
    \item If \emph{yes}, roll two dice to obtain $r$. If $r \leq l$, set $s_h = s_h - 1$. If $r > l$, set $s_h = s_h - 3$. For either outcome, set $l = l - 1$. Go to 6.
    \item If \emph{no}, set $s_h = s_h - 2$. Go to 6.
  \end{enumerate}
  \item If $s_h > 0$ and $s_o \leq 0$, the player has won; if $s_o > 0$ and $s_h \leq 0$, the opponent has won. Otherwise go to 1.
\end{enumerate}

\vspace{0.2cm} It will readily be seen that these dynamics doubly bias battles in favour of the player. First, the opponent has no opportunity to use \igjluck{} to their benefit. Second, when used offensively (to support an attack round that the player has won), the potential benefit (an additional 2 damage; Step 4a in Algorithm 1) outweighs the potential detriment (a reduction of 1 damage; Step 5a in Algorithm 1). When used defensively, this second bias is absent.

We will retain the \igjskill{}, \igjstamina, \igjluck{} terminology throughout this analysis. However, the concepts here can readily be generalised. A player's \igjskill{} can be interpreted as their propensity to be successful in any given competitive round. Their \igjstamina{} can be interpreted as the amount of competitive losses they can suffer before failure. \igjluck{} is the resource that a player can dynamically allocate to gamble on influencing the outcome of competitive rounds, that diminishes with use (so that the success probability of the gamble also diminishes). For example, within the (counter)espionage analogy above, \igjskill{} could be perceived as a company's level of information security, \igjstamina{} its potential to lose sensitive information before failure, and \igjluck{} the resource that it can invest in (counter)espionage.

\section*{Analysis}
In basic combat dynamics, \igjskill{} does not change throughout a battle. The probabilities $p_w, p_d, p_l$ of winning, drawing, an losing an attack round depend only on \igjskill{}, and remain constant throughout the battle. We therefore consider $P^t(s_h, s_o, l)$: the probability, at round $t$, of being in a state where the player has \igjstamina{} $s_h$ and \igjluck{} $l$, and the opponent has \igjstamina{} $s_o$. A discrete-time master equation can readily be written down describing the dynamics above:

  \begin{align}
 &  P^{t+1}(s_h, s_o, l) = \nonumber \\
 & \,\,  p_w (1 - \lambda_1(s_h, s_o+2, l)) P^{t}(s_h, s_o+2, l) + p_w \lambda_1(s_h, s_o+4, l+1) q(l+1) P^{t}(s_h, s_o+4, l+1) \nonumber \\
 & \,\,+ p_w \lambda_1(s_h, s_o+1, l+1) (1-q(l+1)) P^{t}(s_h, s_o+1, l+1) + p_l (1 - \lambda_0(s_h+2, s_o, l)) P^{t}(s_h+2, s_o, l) \nonumber \\
 & \,\,+ p_l \lambda_0(s_h+3, s_o, l+1) (1-q(l+1)) P^{t}(s_h+3, s_o, l+1) + p_l \lambda_0(s_h+1, s_o, l+1) q(l+1) P^{t}(s_h+1, s_o, l+1) \nonumber \\
 & \,\,- (1-p_d) P^{t}(s_h, s_o, l). \label{eqn1} 
\end{align}

Here, $\lambda_i(s_h, s_o, l)$ reflects the decision whether to use \igjluck{} after a given attack round outcome $i$ (player loss $i = 0$ or player win $i = 1$), given the current state of the battle. This decision is free for the player to choose. $q(l)$ and $(1-q(l))$ are respectively the probabilities of a successful and an unsuccessful test against \igjluck{} $l$. We proceed by first considering the case $\lambda_0 = \lambda_1 = 0$ (no use of \igjluck{}). We then consider a simplified case of constant \igjluck{}, before using a dynamic programming approach to address the full general dynamics.

We are concerned with the \emph{victory probability} $v_p$ with which the player is eventually victorious, corresponding to a state where $s_h \geq 1$ and $s_o \leq 0$. We thus consider the `getting to a set' outcome class of this stochastic game \cite{maitra2012discrete}, corresponding to a `winner takes all' race \cite{canbolat2012stochastic}. We start with the probability of winning, drawing, or losing a given round. Let $k_h$ be the player's \igjskill{} and $k_o$ be the opponent's \igjskill{}. Let $D_1$, ..., $D_4$ be random variables drawn from a discrete uniform distribution on $[1, 6]$, and $X = D_1 + D_2 - D_3 - D_4$. Let $\Delta_k = k_h - k_o$. Then, the probabilities of win, draw, and loss events correspond respectively to $p_w = P(X + \Delta_k > 0), p_d = P(X + \Delta_k = 0), p_l = P(X + \Delta_k < 0)$.

$X$ is a discrete distribution on $[-10, 10]$. The point density $P(X = i)$ is proportional to the $(i+10)$th sextinomial coefficient, defined via the generating function $\left( \sum_{j = 0}^5 x^j \right)^n$, for $n = 5$ (specific values $\frac{1}{1296} \{1,\allowbreak 4,\allowbreak 10,\allowbreak 20,\allowbreak 35,\allowbreak 56,\allowbreak 80,\allowbreak 104,\allowbreak 125,\allowbreak 140,\allowbreak 146,\allowbreak 140,\allowbreak 125,\allowbreak 104,\allowbreak 80,\allowbreak 56,\allowbreak 35,\allowbreak 20,\allowbreak 10,\allowbreak 4,\allowbreak 1\}$; OEIS sequence A063260 \cite{oeis}). Hence

\begin{align}
  p_w(\Delta_k) & =  \sum_{j = -\Delta_k+1}^{10} P (X = j) \\
  p_d(\Delta_k) & =  P (X = -\Delta_k) \\
  p_l(\Delta_k) & =  \sum_{j = -10}^{-\Delta_k-1} P(X = j)
\end{align}




\subsection*{Dynamics without luck-based control}
We first consider the straightforward case where the player employs no strategy, never electing to use \igjluck{}. We can then ignore $l$ and consider steps through the $(s_h, s_o)$ \igjstamina{} space, which form a discrete-time Markov chain. Eqn. \ref{eqn1} then becomes

\begin{align}
  P^{t+1}(s_h, s_o, l)  =  p_w P^t(s_h, s_o+2, l) + p_l P^t(s_h+2, s_o, l)  - (1-p_d) P^t(s_h, s_o, l). \label{eqn2}
\end{align}

We can consider a combinatorial approach based on `game histories' describing steps moving through this space \cite{maitra2012discrete}. Here a game history is a string from the alphabet $\{W, D, L\}$, with the character at a given position $i$ corresponding to respectively to a win, draw, loss in round $i$. We aim to enumerate the number of possible game histories that correspond to a given outcome, and assign each a probability.

We write $w, d, l$ for the character counts of $W, D, L$ in a given game history. A victorious game must always end in $W$. Consider the string describing a game history omitting this final $W$. First leaving out $D$s, we have $(w-1)$ $W$s and $l$ $L$s that can be arranged in any order. We therefore have $n(w,l) = \binom{w - 1 + l}{l}$ possible strings, each of length $w-1+l$. For completeness, we can then place any number $d$ of $D$s within these strings, obtaining

\begin{equation}
  n(w,l,d) = \binom{w - 1 + l}{l} \binom{w - 1 + l + d}{d}.
\end{equation}

Write $\sigma_h = \lfloor s_h/2 \rfloor$ and $\sigma_o = \lfloor s_o/2 \rfloor$, describing the number of rounds each character can lose before dying. Then, for a player victory, $w = \sigma_o$ and $l \leq \sigma_h - 1$. $d$ can take any nonnegative integer value. The appearance of each character in a game string is accompanied by a multiplicative factor of the corresponding probability, so we obtain

\begin{equation}
  v_p = p_w^{\sigma_o} \sum_{l=0}^{\sigma_h-1} \sum_{d = 0}^{\infty} p_l^l p_d^d \binom{w - 1 + l}{l} \binom{w - 1 + l + d}{d},
\end{equation}

where the probability associated with the final $W$ character has now also been included. After some algebra, and writing $\rho_w = p_w / (1-p_d)$ and $\rho_l = p_l / (1-p_d)$, this expression becomes





\begin{equation}
  v_p =  p_w^{\sigma_o} (p_l + p_w)^{-\sigma_o} \Bigg( \rho_w^{-\sigma_o} - p_l^{\sigma_h} (p_l + p_w)^{-\sigma_h}  \times \binom{\sigma_h + \sigma_o - 1}{\sigma_h} \, {}_2 F_1 \left( 1, \sigma_h + \sigma_o, \sigma_h + 1, \rho_l \right) \Bigg)
  \label{basiceqn}
  \end{equation}

\begin{figure}
  \centering
  \includegraphics[width=8cm]{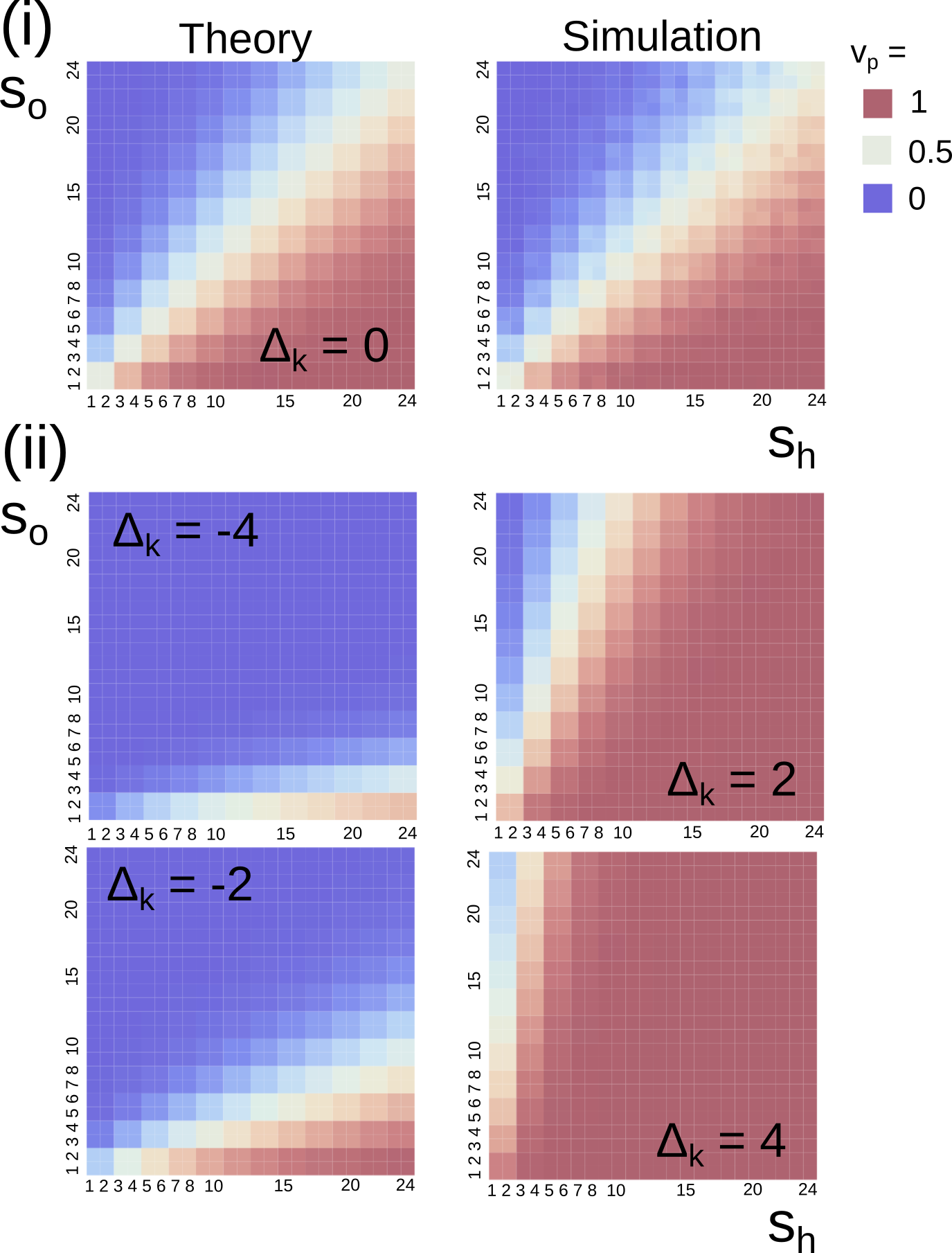}
  \caption{\footnotesize \textbf{Victory probability in the absence of \igjluck{}-based strategy.} (i) Comparison of predicted victory probability $v_p$ from Eqn. \ref{basiceqn} with stochastic simulation. (ii) $v_p$ behaviour as \igjskill{} difference $\Delta_k$ changes.}
  \label{basiccompare}
\end{figure}

Eqn. \ref{basiceqn} is compared with the result of stochastic simulation in Fig. \ref{basiccompare}, and shown for various \igjskill{} differences $\Delta_k$ and $(s_h, s_o)$ initial conditions. Intuitively, more favourable $\Delta_k > 0$ increase $v_p$ and less favourable $\Delta_k < 0$ decrease $v_p$ for any given state, and discrepancies between starting $s_h$ and $s_o$ also influence eventual $v_p$. A pronounced $s \Mod{2}$ structure is observed, as in the absence of \igjluck{}, $s = 2n$ is functionally equivalent to $s = 2n-1$ for integer $n$. For lower initial \igjstamina s, $v_p$ distributions become more sharply peaked with $s$ values, as fewer events are required for an eventual outcome.

\subsection*{Analytic dynamics with simplified \igjluck{}-based control}

To increase the probability of victory beyond the basic case in Eqn. \ref{basiceqn}, the player can elect to use \igjluck{} in any given round. We will first demonstrate that the above history-counting analysis can obtain analytic results when applied to a simplified situation when \igjluck{} is not depleted by use, so that the only limit on its employment is its initial level \cite{blanning1981variable}. We then employ a dynamic programming scheme to analyse the more involved general case.

First consider the case where \igjluck{} is only used offensively (step 4a in Algorithm 1), and is used in every successful attack round. For now, ignore losses and draws. Then every game history consists of $A$s and $B$s, where $A$ is a successful offensive use of \igjluck{} and $B$ is an unsuccessful offensive use of \igjluck{}.

Consider the game histories that lead to the opponent losing exactly $n$ \igjstamina{} points \emph{before} the final victorious round. There are $\lfloor n/4 \rfloor + 1$ string lengths that can achieve this, which are $L = n - 3k$, where $k$ runs from $0$ to $\lfloor n/3 \rfloor$. The strings with a given $k$ involve $n-4k$ failures and $k$ successes.

If we make the simplifying assumption that \igjluck{} is not depleted with use, every outcome of a \igjluck{} test has the same success probability $q(l) = q$. Then the problem is simplified to finding the number of ways of arranging $k$ $A$s and $(n-4k)$ $B$s for each possible string:


\begin{equation}
  N(k) = \frac{(n-3k)!}{k! (n-4k)!} 
\end{equation}

Now, for every string with a given $k$, with corresponding string length $n-3k$, we can place $l$ $L$s and $d$ $D$s as before, giving

\begin{equation}
N(k; n, l, d) = \frac{(n-3k)!}{k! (n-4k)!} \binom{n-3k + l}{l} \binom{n-3k + l + d}{d}.
\end{equation}

The complete history involves a final victorious round. For now we will write the probability of this event as $p_f$, then the probability associated with this set of histories is

\begin{align}
  P(k; n, l, d, p_f)  = p_f \frac{(n-3k)!}{k! (n-4k)!} q^k (1-q)^{n-4k}  p_l^l \binom{n-3k + l}{l} p_d^d \binom{n-3k + l + d}{d} \label{stringcount}
\end{align}

and, as $P(k; n, p_f) = \sum_{l=0}^{\sigma_{h}-1} \sum_{d=0}^{\infty} P(k; n, l, d, p_f)$,

\begin{align}
  P(k; n, p_f)  = &  p_f \frac{1}{k!(n-4k)!} (1-p_d)^{3k-n-2-\sigma_{h}} (p_w-p_w q)^{n-4k} (p_w q)^k \rho^{-n-1} (n-3k)! \nonumber \\
  & \times \bigg[ (1-p_d)^{\sigma_h+1} \rho^{3k} + p_l^{\sigma_h} (p_l+p_d-1) \rho^n \binom{n-3k+\sigma_h}{\sigma_h} {}_2F_1(1, 1-3k+n+\sigma_h, 1+\sigma_h, p_l/(1-p_d)) \bigg]
\end{align}

where $\rho = (p_d+p_l-1)/(p_d-1)$. Hence

\begin{equation}
  P_n(n, p_f) = \sum_{k=0}^{n/3} P(k; n, p_f)
\end{equation}

Now consider the different forms that the final victorious round can take. The opponent's \igjstamina{} can be reduced to 4 followed by an $A$, 3 followed by $A$, 2 followed by $A$, or 1 followed by $A$ or $B$. If we write $P(m; X)$ for the probability of reducing the opponent's \igjstamina{} to $m$ then finishing with event $X$,

\begin{equation}
  v_p = P(4; A)+P(3;A)+P(2;A)+P(1;A)+P(1;B)
\end{equation}

hence

\begin{align}
  v_p =  P_n(s_o-4, q)+P_n(s_o-3, q)+P_n(s_o-2, q) +P_n(s_o-1, q)+P_n(s_o-1, (1-q)). \label{dumbluckeqn}
\end{align}

\begin{figure}
  \centering
  \includegraphics[width=8cm]{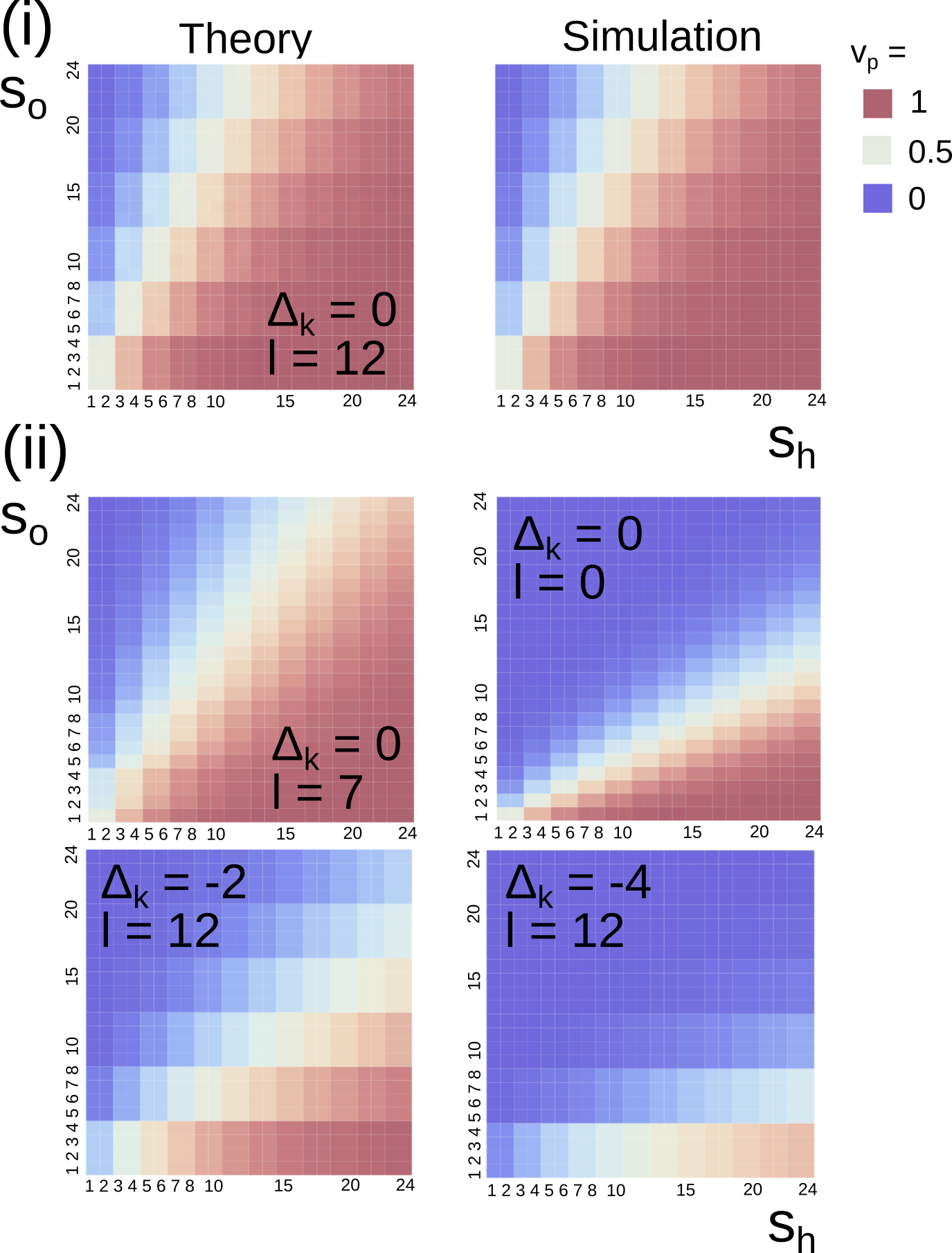}
  \caption{\footnotesize \textbf{Victory probability with constant \igjluck{} employed offensively.} (i) Eqn. \ref{dumbluckeqn} compared to stochastic simulation. (ii) Constant luck outcomes for intermediate and low \igjluck{}, and for high \igjluck{} mitigating low $\Delta_k$ values.}
  \label{dumbluck}
\end{figure}

Fig. \ref{dumbluck}(i) compares Eqn. \ref{dumbluckeqn} and stochastic simulation, and shows that use of \igjluck{} can dramatically increase victory probability in a range of circumstances. Similar expressions can be derived for the defensive case, where \igjluck{} is solely used when a round is lost, and with some relaxations on the structure of the sums involved the case where \igjluck{} is not used in every round can also be considered.

Clearly, for constant \igjluck{}, the optimal strategy is to always use \igjluck{} when the expected outcome is positive, and never when it is negative. The expected outcomes for offensive and defensive strategies can readily be computed. For offensive use, the expected damage dealt when using \igjluck{} is

\begin{equation}
  \langle d \rangle = 1 \times (1-q) + 4 \times q \label{leqn1}
\end{equation}

In the absence of \igjluck{}, $d = 2$ damage is dealt, so the expected outcome using \igjluck{} is beneficial if $\langle d \rangle > 2$. We therefore obtain $q > \frac{1}{3}$ as the criterion for employing \igjluck{}. For comparison, the probability of scoring 6 or under on two dice is $q(6) = 0.417$ and the probability of scoring 5 or under on two dice is $q(5) = 0.278$.

For defensive use, the expected damage received when using \igjluck{} in an attack round is

\begin{equation}
  \langle d \rangle = 1 \times q + 3 \times (1-q) \label{leqn2}
\end{equation}

In the absence of \igjluck{}, $d = 2$ damage is taken, and we thus now obtain $q > \frac{1}{2}$ as our criterion. For comparison, the probability of scoring 7 or under on two dice is $q(7) = 0.583$.

The counting-based analyses above assume that \igjluck{} stays constant. The dynamics of the game actually lead to \igjluck{} diminishing every time is it is used. Instead of the outcome of a \igjluck{} test being a constant $q$, it now becomes a function of how many tests have occurred previously.

To explore this situation, consider the above case where \igjluck{} is always used offensively and never defensively. Describe a given string of \igjluck{} outcomes as an ordered set $\mathcal{V}$ of indices labelling where in a string successes occur. For example, the string of length $L = 5$ with $\mathcal{V} = {2, 4}$ would be $BABAB$. Let $\mathcal{V}' = \mathcal{V}^c \setminus \{1, ..., L\}$. Then

\begin{equation}
  P(\mathcal{V}) = \prod_{i \in \mathcal{V}} q(i) \prod_{i \in \mathcal{V}'} (1-q(i))
\end{equation}

where $q(i)$ is the probability of success on the $i$th \igjluck{} test.

As above, for a given $k$, $L = n-3k$, and we let $\mathcal{S}(k)$ be the set of ordered sets with $k$ different elements between 1 and $n-3k$. Then the probability of a given string of $A$s and $B$s arising is

\begin{equation}
  P(n) = \sum_{k=0}^{n/3} \sum_{\mathcal{V} \in \mathcal{S}(k)} \prod_{i \in \mathcal{V}} q(i) \prod_{i \in \mathcal{V}'} (1-q(i))
\end{equation}

We can no longer use the simple counting argument in Eqn. \ref{stringcount} to compute the probabilities of each history, because each probability now depends on the specific structure of the history. It will be possible to enumerate these histories exhaustively but the analysis rapidly expands beyond the point of useful interpretation, so we turn to dynamic programming to investigate the system's behaviour.

\subsection*{Stochastic optimal control with dynamic programming}

In a game with a given $\Delta_k$, we characterise every state of the system with a tuple $\mathcal{S} = \{s_h, s_o, l, O\}$ where $O$ is the outcome (win or loss) of the current attack round. The question is, given a state, should the player elect to use \igjluck{} or not?

A common approach to identify the optimal strategy for a Markov decision problem in a discrete state space is to use the Bellman equation \cite{bellman1957markovian, kirk2012optimal}, which in our case is simply

\begin{equation}
  v_p(\mathcal{S}) = \max_a \left( \sum_{\mathcal{S}'} P_a (\mathcal{S}, \mathcal{S}') v_p (\mathcal{S}') \right).
\end{equation}


Here, $a$ is a strategy dictating what action to take in state $\mathcal{S}$, $P_a(\mathcal{S}, \mathcal{S}')$ is the probability under strategy $a$ of the transition from state $\mathcal{S}$ to state $\mathcal{S}'$, and $v_p(\mathcal{S})$ is the probability-to-victory of state $\mathcal{S}$. The joint problem is to compute the optimal $v_p$, and the strategy $a$ that maximises it, for all states. To do so, we employ a dynamic programming approach of backward induction \cite{bellman1957markovian}, starting from states where $v_p$ is known and computing backwards through potential precursor states.


\begin{figure}
  \includegraphics[width=12cm]{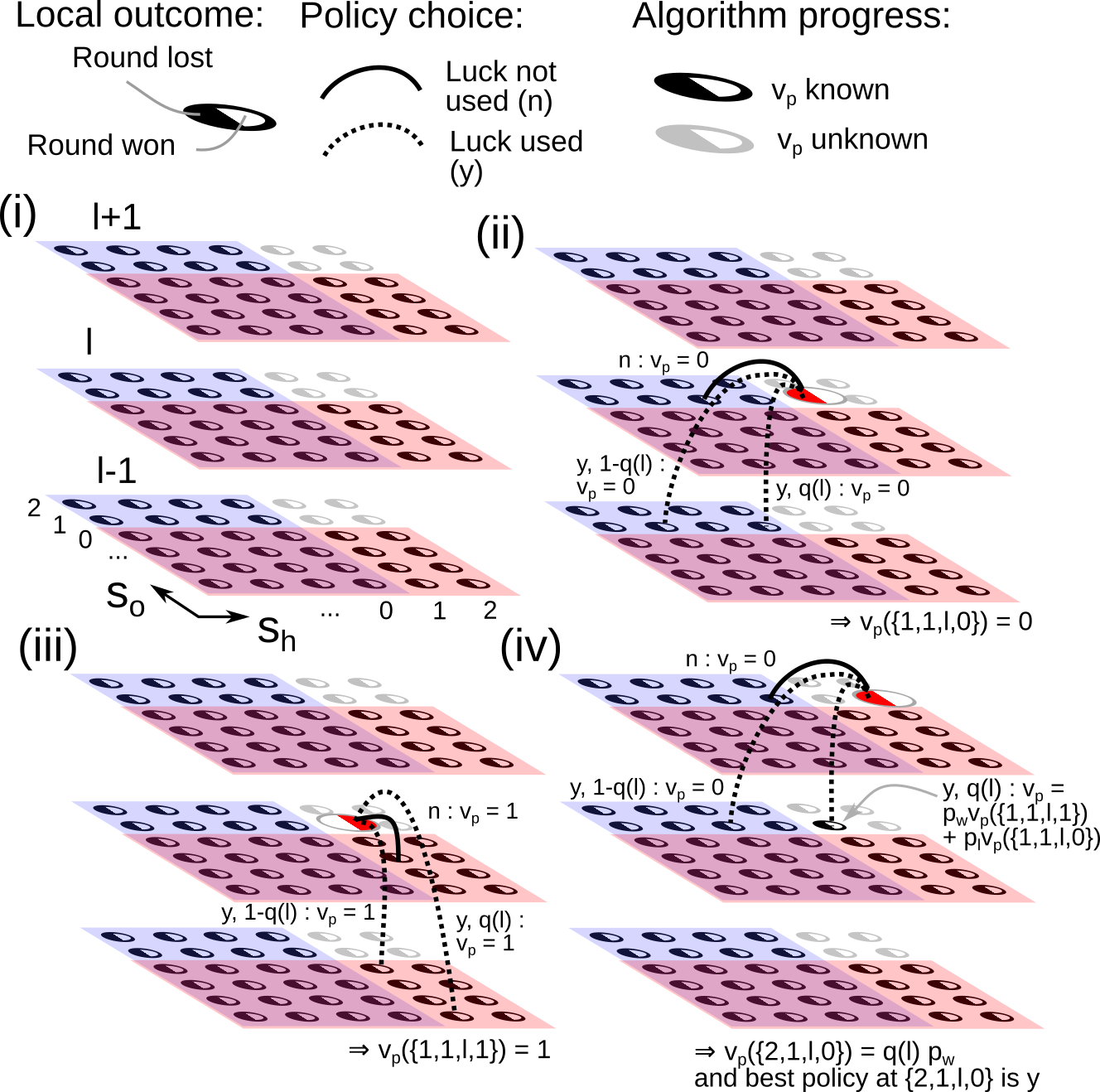}
  \caption{\footnotesize \textbf{Dynamic programming scheme.} The discrete state space of the game is represented by horizontal ($s_h$) and vertical ($s_o$) axes, layers ($l$) and a binary outcome at each position ($O$). (i) First, $v_p = 0$ is assigned to states corresponding to loss ($s_h \leq 0$, blue) and $v_p = 1$ is assigned to states corresponding to victory ($s_o \leq 0$, red). Then, iteratively, all states where all outcomes have an assigned $v_p$ are considered. (ii) Considering outcomes after a loss in the current state (here, ${1, 1, l, 0}$, highlighted red). The probability-to-victory for a `no' strategy corresponds to the solid line; the probability-to-victory for a `yes' strategy consists of contributions from both dashed lines (both outcomes from a \igjluck{} test). (iii) Same process for a win outcome from the current state. (iv) Now the previous node has characterised probabilities-to-victory for both loss and win outcomes, a next iteration of nodes can be considered. Now, the probability-to-victory from a `yes' strategy has a term corresponding to the potential outcomes from the previously characterised node.}
  \label{dyncartoon} 
\end{figure}

The dynamic programming approach first assigns a probability-to-victory $v_p$ for termination states (Fig. \ref{dyncartoon}(i)). For all `defeat' states with $s_h \leq 0, s_o > 0$, we set $v_p = 0$; for all `victory' states with $s_o \leq 0, s_h > 0$, set $v_p = 1$ (states where both $s_h \leq 0$ and $s_o \leq 0$ are inaccessible). We then iteratively consider all states in the system where $v_p$ is fully determined for a loss outcome from the current round (Fig. \ref{dyncartoon}(ii)), and similarly for a win outcome (Fig. \ref{dyncartoon}(iii)).

For states involving a loss outcome, we compute two probability propagators. The first corresponds to the strategy where the player elects to use \igjluck{}, and is of magnitude

\begin{align}
  p_y  =  &\sum_{\mathcal{S}'} P_y (\mathcal{S}, \mathcal{S}') v_p(\mathcal{S}') \nonumber \\
  = & \, q(l) p_l v_p(\{s_h-1, s_o, l-1, 0\}) + (1-q(l)) p_l v_p(\{s_h-3, s_o, l-1, 0\}) \nonumber \\
 & + \, q(l) p_w v_p(\{s_h-1, s_o, l-1, 1\}) + (1-q(l)) p_w v_p(\{s_h-3, s_o, l-1, 1\})
\end{align}

the second corresponds to the strategy where the player does not use \igjluck{}, and is 

\begin{align}
  p_n  = & \, \sum_{\mathcal{S}'} P_n (\mathcal{S}, \mathcal{S}') v_p(\mathcal{S}') \nonumber \\
  = & \,p_l v_p(\{s_h-2, s_o, l, 0\}) + p_w v_p(\{s_h-2, s_o, l, 1\}).
\end{align}

When considering the probability-to-victory for a given state, we hence consider both the next $(s_h, s_o, l)$ combination that a given event will lead to, and also both possible outcomes (win or loss) from this state. For a given state, if $p_y > p_n$, we record the optimal strategy as using \igjluck{} and record $v_p = p_y$; otherwise we record the optimal strategy as not to use \igjluck{} and record $v_p = p_n$. In practise we replace $p_y > p_n$ with the condition $p_y > (1 + 10^{-10}) p_n$ to avoid numerical artefacts, thus requiring that the use of \igjluck{} has a \emph{relative} advantage to $v_p$ above $10^{-10}$.

We do the same for states involving a win outcome from this round, where the two probability propagators are now

\begin{align}
  p_y  = & \, q(l) p_l v_p(\{s_h, s_o-4, l-1, 0\}) + (1-q(l)) p_l v_p(\{s_h, s_o-1, l-1, 0\}) \nonumber \\
 & +\, q(l) p_w v_p(\{s_h, s_o-4, l-1, 1\}) + (1-q(l)) p_w v_p(\{s_h, s_o-1, l-1, 1\});
\end{align}

and 

\begin{align}
  p_n = & \,p_l v_p(\{s_h, s_o-2, l, 0\}) + p_w v_p(\{s_h, s_o-2, l, 1\}).
\end{align}

\begin{figure*}
  \centering
    \includegraphics[width=17cm]{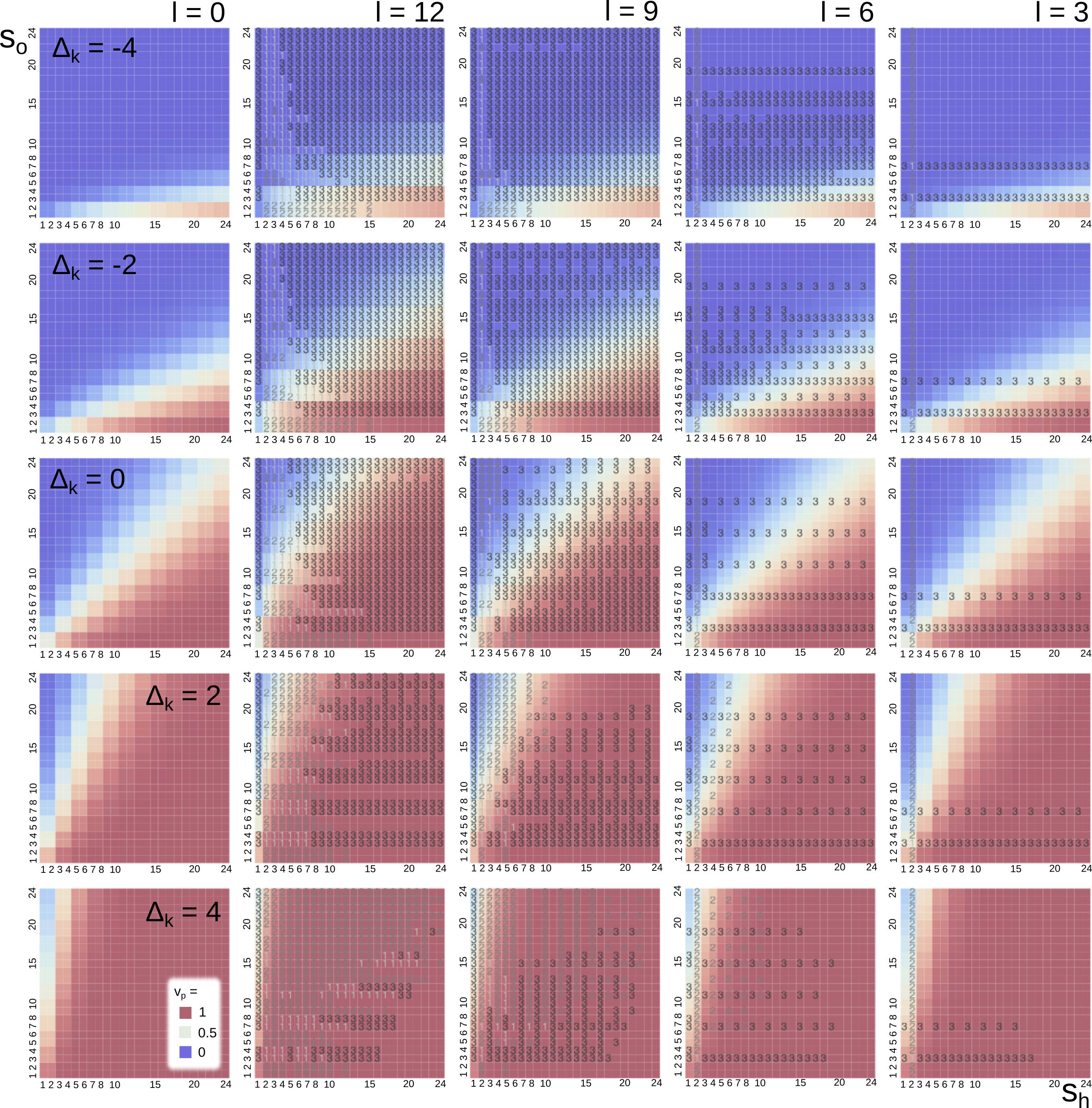}
  \caption{\footnotesize \textbf{Optimal strategies and victory probabilities throughout state space.} The optimal strategy at a given state is given by the number at the corresponding point (1 -- use \igjluck{} regardless of the outcome of this round; 2 -- use \igjluck{} if this round is lost; 3 -- use \igjluck{} if this round is won). No number means that the optimal strategy is not to employ \igjluck{} regardless of outcome. Colour gives victory probability $v_p$.}
  \label{fullluckdyn-analytic}
\end{figure*}

Each new pair of states for which the optimal $v_p$ is calculated opens up the opportunity to compute $v_p$ for new pairs of states (Fig. \ref{dyncartoon}(iv)). Eventually a $v_p$ and optimal strategy is computed for each outcome, providing a full `roadmap' of the optimal decision to make under any circumstance. This full map is shown in Supplementary Fig. \ref{suppfig}, with a subset of states shown in Fig. \ref{fullluckdyn-analytic}.

\begin{figure*}
  \centering
    \includegraphics[width=17cm]{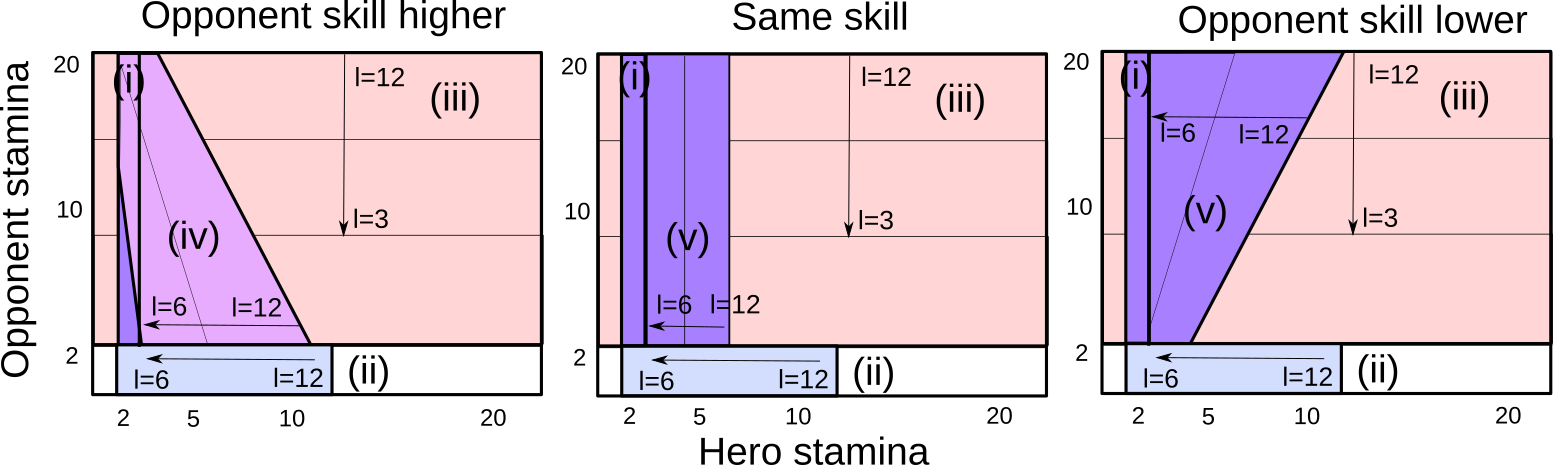}
  \caption{\footnotesize \textbf{Summary of optimal strategies and victory probabilities throughout state space.} Arrows depict regions where the optimal policy depends on current available \igjluck{}, including the general change in that region with decreasing \igjluck{}. Empty regions are those where avoiding the use of \igjluck{} is the optimal policy. Other policies: (i) Hero \igjstamina{} $= 2$: use luck defensively (but see (iv), (v)); (ii) Opponent \igjstamina{} $\leq 2$: use luck defensively; (iii) Use \igjluck{} offensively, particularly when opponent \igjstamina{} $\Mod 4 = 3$; (iv) Use \igjluck{} offensively and defensively; (v) Use \igjluck{} defensively, and offensively if luck is high and opponent \igjstamina{} $\Mod 4 = 3$ or $0$.}
  \label{fullluckdyn-illus}
\end{figure*}


We find that a high \igjluck{} score and judicious use of \igjluck{} can dramatically enhance victory probability against some opponents. As an extreme example, with a \igjskill{} detriment of $\Delta_k = -9$, $s_h = 2, s_o = 23, l = 12$, use of \igjluck{} increased victory probability by a factor of $10^{18}$, albeit to a mere $2.3 \times 10^{-19}$. On more reasonable scales, with a \igjskill{} detriment of $\Delta_k = -4$, $s_h = 22, s_o = 19, l = 12$, use of \igjluck{} increased victory probability 1159-fold, from $9.8 \times 10^{-6}$ to 0.011 (the highest fold increase with the final probability greater than 0.01). With a \igjskill{} detriment of $\Delta_k = -2$, $s_h = 22, s_o = 21, l = 12$, use of \igjluck{} increased victory probability 21-fold, from 0.010 to 0.22 (the highest fold increase with the initial probability greater than 0.01). Using encounters that appear in the FF universe \cite{outofthepit}, for a player with maximum statistics $k_h = 12, s_h = 24, l = 12$, optimal use of \igjluck{} makes victory against an adult White Dragon ($k_o = 15, s_o = 22$) merely quite unlikely ($v_p = 0.046$) rather than implausible ($v_p = 4.4 \times 10^{-4}$), and victory against a Hell Demon ($k_o = 14, s_o = 12$) fairly straightforward ($v_p = 0.78$) rather than unlikely ($v_p = 0.28$).

\subsubsection*{Structure of optimal policy space}
  
There is substantial similarity in optimal policy choice between several regions of state space. For large \igjskill{} deficiencies $\Delta_k \leq -6$ (low probability of victory) the distribution of optimal strategies in \igjstamina{} space is the same for a given $l$ for all $\Delta_k$. For higher $l$, this similarity continues to higher $\Delta_k$; for $l = 12$, only 5 points in \igjstamina{} space have different optimal strategies for $\Delta_k = -9$ and $\Delta_k = -4$. At more reasonable victory probabilities, a moderate transition is apparent between $l = 6$ and $l = 5$, where the number of points in \igjstamina{} space where the optimal strategy involves using \igjluck{} decreases noticeably. This reflects the lower expected advantage for $l=5$ (Eqns. \ref{leqn1}-\ref{leqn2}).

The interplay of several general strategies is observed in the optimal structures. First note that $\lceil s / 4 \rceil$ gives the number of hits required for defeat if a hit takes 4 \igjstamina{} points (a successful offensive \igjluck{} test) and $\lceil s / 2 \rceil$ gives the number of hits required for defeat in the absence of strategy. These scales partition \igjstamina{} space by the number of rounds required for a given outcome and hence dictate several of the `banded' structures observable in strategy structure. For example, at $s_h = 2$, it is very advantageous for the player to attempt to mitigate the effect of losing another round. Almost all circumstances display a band of defensive optimal strategy at $s_h = 2$. 

At $s_o = 3$, a successful offensive \igjluck{} test is very advantageous (immediate victory). An unsuccessful offensive test, leading to $s_o = 2$, is not disadvantageous to the same extent: we still need exactly one successful attack round without \igjluck{}, as we would if we had not used \igjluck{} and achieved $s_o = 1$ instead. A strip of strategy 3 (or strategy 1) at $s_o = 3$ is thus the next most robustly observed feature, disappearing only when victory probability is already overwhelmingly high. Many other structural features result from a tradeoff between conserving \igjluck{} and increasing the probability of encountering this advantageous region. An illustration of the broad layout of optimal strategies is shown in Fig. \ref{fullluckdyn-illus}; a more fine-grained analysis is provided in the Appendix.

\subsection*{Simulated dynamics with heuristic \igjluck{} strategies}

While the dynamic programming approach above gives the optimal strategy for any circumstance, the detailed information involved does not lend itself to easy memorisation. As in Smith's discussion of solitaire, `the curse of dimensionality applies for the state description, and most wise players use heuristics' \cite{smith2007dynamic}. We therefore consider, in addition to the semi-quantitative summary in Fig. \ref{fullluckdyn-illus}, coarse-grained quantitative `strategies' that, rather than specifying an action for each branch of the possible tree of circumstances, use a heuristic rule that is applied in all circumstances. 

Among the simplest strategies are `open-loop' rules, where the current state of the player's and the opponent's \igjstamina{} does not inform the choice of whether to use \igjluck{}. We may also consider `feedback' rules, where the current state of \igjstamina s informs this choice. We parameterise \igjluck{} strategies by a threshold $\tau$; \igjluck{} will only be used if the player's current \igjluck{} score is greater than or equal to $\tau$.

We consider the following \igjluck{} strategies:
\begin{enumerate}[noitemsep]
  \setcounter{enumi}{-1}
\item Never use \igjluck{} ($\lambda_0 = \lambda_1 = 0$ as above);
\item Use \igjluck{} in every non-draw round (i.e. to both ameliorate damage and enhance hits, $\lambda_0 = \lambda_1 = \Theta(l > \tau)$);
\item Use \igjluck{} defensively (i.e. only to ameliorate damage, $\lambda_1 = 0, \lambda_0 = \Theta(l > \tau)$);
\item Use \igjluck{} offensively (i.e. only to enhance hits, $\lambda_0 = 0, \lambda_1 = \Theta(l > \tau)$);
\item Use \igjluck{} defensively if $l > \tau$ and $s_h < 6$;
\item Use \igjluck{} offensively if $l > \tau$ and $s_o > 6$;
\item Use \igjluck{} if $l > \tau$ and $s_h < 6$;
\item Use \igjluck{} if $l > \tau$ and $s_h < 4$;
  \item Use \igjluck{} if $l > \tau$ and $s_h < s_o$.
\end{enumerate}
Strategy 0 is the trivial absence of \igjluck{} use; strategies 1-3 are `open-loop' and 4-8 are `feedback'. Strategies 2 and 4 are defensive; strategies 3 and 5 are offensive; strategies 1, 6, 7, 8 are both offensive and defensive.



\begin{figure*}
  \centering
      \includegraphics[width=17cm]{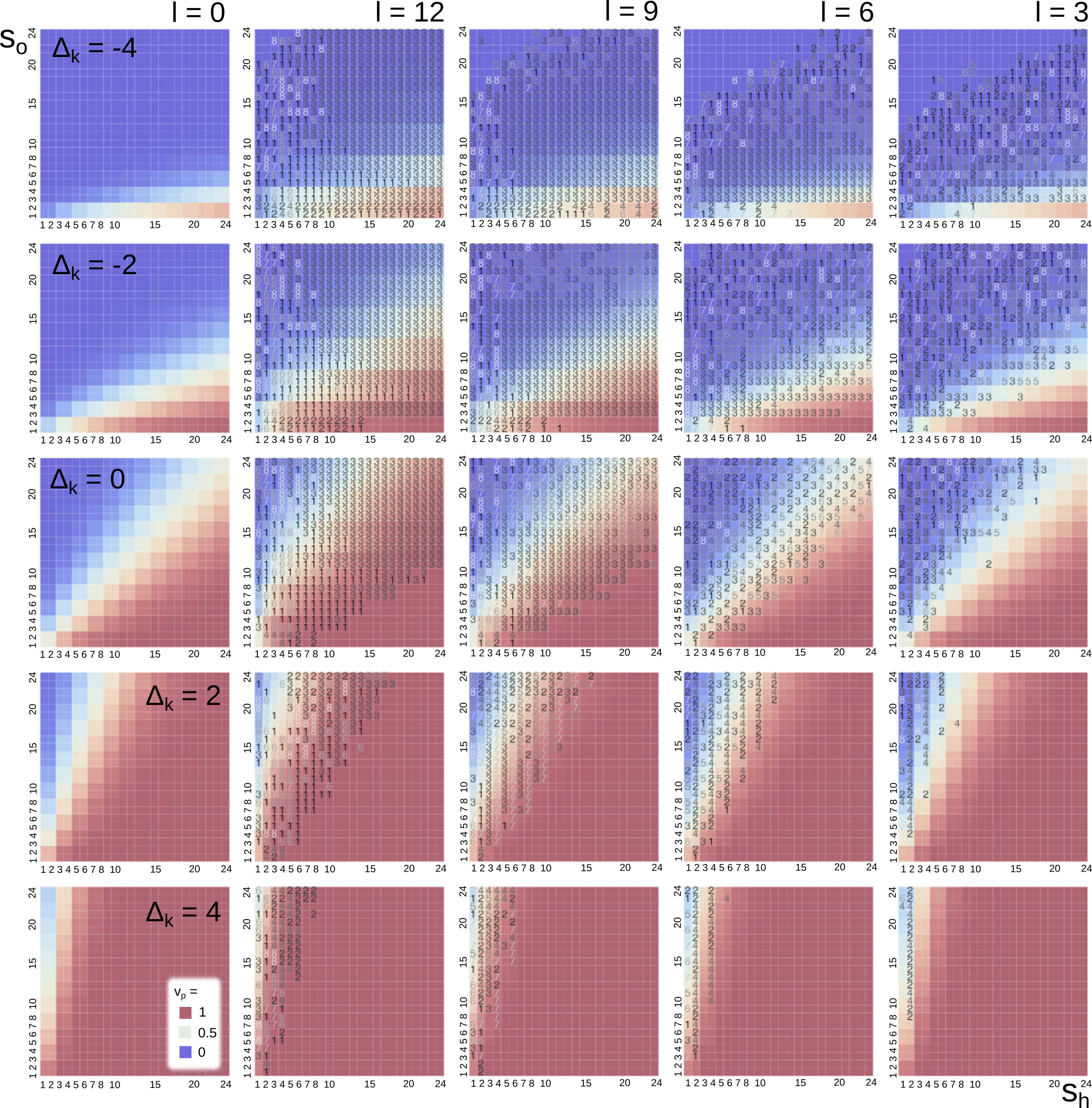}
  \caption{\footnotesize \textbf{Simulated heuristic successful strategies and victory probabilities throughout state space.} Here, a given strategy is used in every round until combat ends. The strategy with the most positive effect on $v_p$ in stochastic simulation for a given state is given by the number at the corresponding point (labels in the main text). No number means that no strategy had a positive influence on $v_p$. Colour gives victory probability $v_p$.}
  \label{fullluckdyn-sim}
\end{figure*}

The analytic structure determined above is largely visible in the simulation results (Supplementary Fig. \ref{fullluckdyn-sim}). The main differences arise from the construction that the same strategy is retained throughout combat in the simulation study. For low $\Delta_k$, high $l$, the favouring of strategy 3 except at low $s_h$ (strategy 1) and $s_o < 2$ (strategy 2) is clear. At low $s_h$, some feedback strategies gain traction: for example, strategy 8, which is identical to strategy 1 when $l$ is high and $s_h < s_o$, but avoids depleting \igjluck{} if ever $s_h \geq s_o$, is more conservative than always using strategy 1. Similarly, strategy 5 provides a feedback alternative to strategy 3, using \igjluck{} only at $s_o > 6$, which is favoured at intermediate $l$ and high $s_o$ where it is beneficial to preserve \igjluck{}. These feedback-controlled options mimic the state-specific application of strategy found in the optimal cases in Fig. \ref{fullluckdyn-analytic}. 

Decreasing $l$ introduces the sparsification of the strategy 3 regions -- an effect which, as with the analysis, is more pronounced at higher $\Delta_k$. In simulations, at higher $\Delta_k$, the increasingly small relative increases in victory probability mean that several regions do not display substantial advantages to using strategy. However, at moderate to positive $\Delta_k$, the increased takeover of strategy 2 is observed for high $s_o$, low $s_h$. Strategy 4, a more conservative feedback version of strategy 2, also experiences support in these regions, competing with strategy 2 particularly in bands where $s_h \Mod{2} = 0$, as described in the analytic case. The above banded structures of strategies 2 and 3, and the ubiquitous strategy 2 (or equivalent) at $s_h = 2$, are also observed for intermediate $l$. 

\begin{figure}
  \includegraphics[width=8cm]{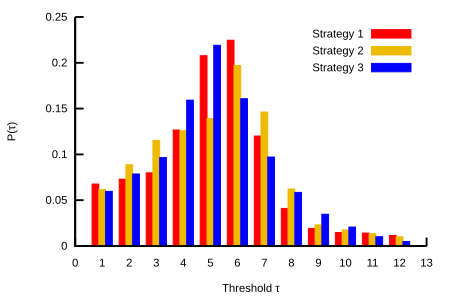}
\caption{\footnotesize \textbf{Probability of \igjluck{} threshold $\tau$ appearing in an optimal strategy.} Histogram shows $\tau$ values appearing in heuristic approaches to optimise $v_p$ in stochastic simulation, featuring strategies 1-3 from the text.}
  \label{taudist}
\end{figure}

Under these simplified rulesets, what is the optimal threshold $\tau$ above which we should use \igjluck{}? Straightforward application of Eqns. \ref{leqn1}-\ref{leqn2} are supported by simulation results (Supplementary Fig. \ref{taudist}), where the probability that a given $\tau$ appears in the highest $v_p$ strategy peaked at $\tau = 6$ for strategy 1 and $\tau = 5$ for strategy 3.

\begin{figure*}
  \centering
  \includegraphics[width=16cm]{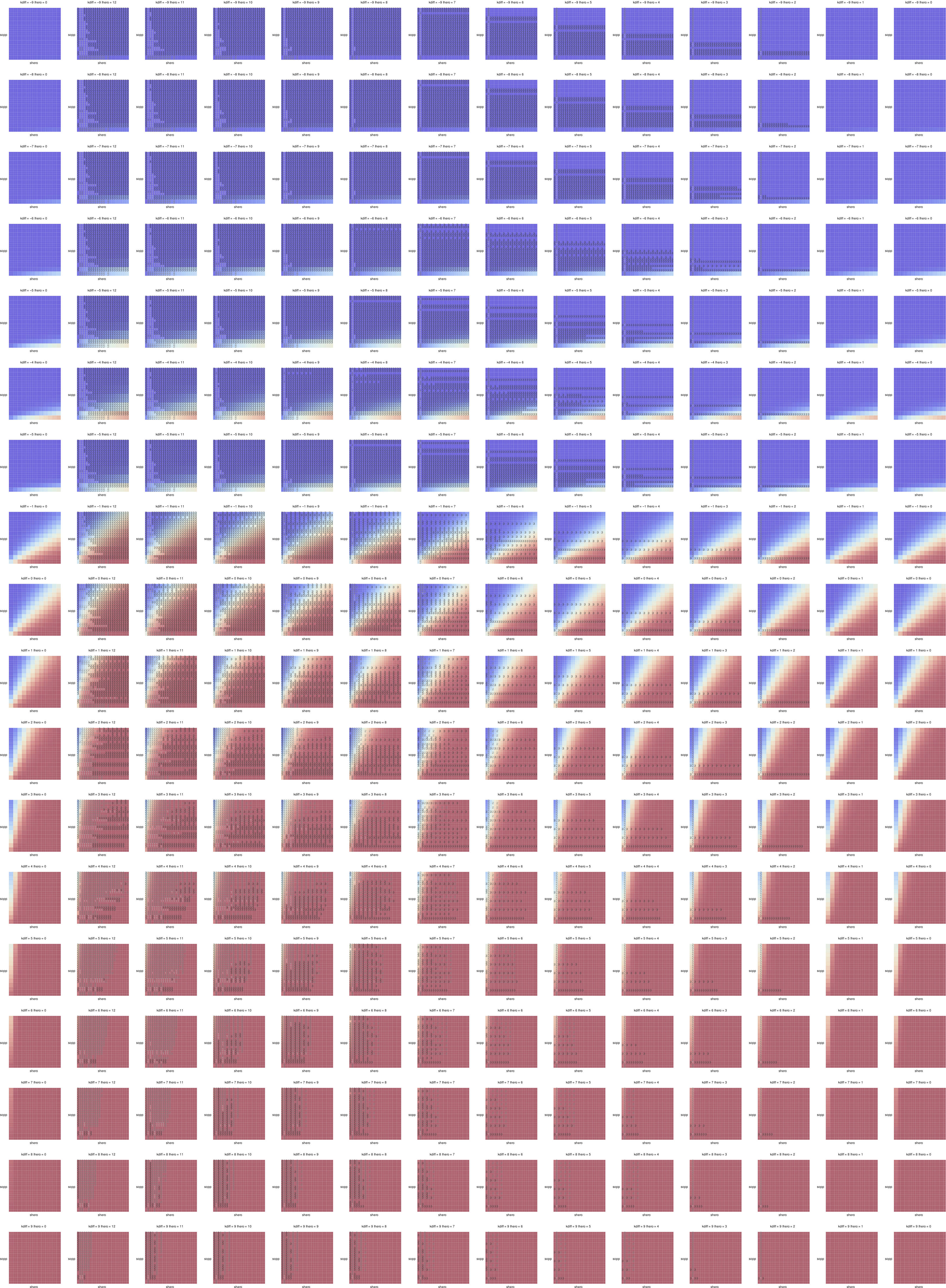}
  \caption{\footnotesize (magnification recommended) \textbf{Complete set of optimal strategies for all states.} The optimal strategy for any given state $s_h$ (shero), $s_o$ (sopp), $l$ (lhero), $\Delta_k$ (kdiff). 1 -- use \igjluck{} regardless of the outcome of this round; 2 -- use \igjluck{} if this round is lost; 3 -- use \igjluck{} if this round is won. No number means that the optimal strategy is not to employ \igjluck{} regardless of outcome. Colour gives victory probability $v_p$.}
  \label{suppfig}
\end{figure*}

A heuristic approach harnessing these insights, at the broad level of Fig. \ref{fullluckdyn-illus}, might be as follows. Employ strategy 3 (offensive \igjluck{} use), unless: (i) initial \igjluck{} is high or $s_o$ is low, in which case strategy 1 (unconditional \igjluck{} use); (ii) $k_h > k_o$ and (i) is not true, in which case strategy 2 (defensive \igjluck{} use); or (iii) $k_o > k_h$ and $s_o \leq 2$ and (i) is not true, in which case strategy 2 (defensive \igjluck{} use). At the next level of detail, feedback control in the form of strategies 4-8 may be beneficial in some cases, notably when \igjluck{} depletion is likely to become an issue; regions where this is likely to be the case can be read off from Supplementary Fig. \ref{fullluckdyn-sim}. Finally, the optimal strategy for any given state can be read off from Supplementary Fig. \ref{suppfig}.




\section*{Discussion}
We have examined the probability of victory in an iterated, probabilistic game that plays a central role in a well-known and widespread interactive fiction series. The game can be played with or without `strategy', here manifest by the consumption of a limited resource to probabilistically influence the outcome of each round.

Several interesting features of the FF combat system make it potentially noteworthy with respect to similar systems \cite{canbolat2012stochastic, neller2004optimal, smith2007dynamic}. The allocation of resource is dynamic and depends on system state \cite{blanning1981variable, hopp1987sequential, posner1990optimal}. The use of resource can both increase the probability of a positive outcome for the player, or a negative outcome for the opponent. Use of this resource does not guarantee a positive outcome: its use is a gamble \cite{maitra2012discrete, dubins1965inequalities} that may negatively affect the player. The probability of this negative effect increases as more resource is used, providing an important consideration in the decision of whether to invoke this policy in the face of `diminishing returns' \cite{deckro2003modeling}.

The fact that gambles are dynamically used to influence iterated stochastic dynamics complicates the analysis of the game. To employ dynamic programming, state space was labelled both by the current statistics of the players and by the outcome of the current stochastic round. The policy decision of whether to gamble thus depends both on current statistics and the outcome of a round.

This analysis reveals several structural properties of the system that are not specific to the FF context of this study. The case of a limited resource being gambled against to both amplify successes and mitigate failures (with the probability of a positive outcome diminishing as resource is used) bears some resemblance to, for example, a conceptual picture of espionage and counterespionage \cite{solan2004games}. For example, resource could be allocated to covert operations with some probability of amplifying thefts and mitigating losses of information, with increased probability of negative outcomes due to discovery as these covert activites are employed more. Our analysis then reveals the strategies to be employed in different scenarios of information security (\igjskill{}) and robustness to information loss (\igjstamina).

In the absence of strategy, we find a closed-form expression for victory probability that takes an intuitive form. When strategy is included, dramatic increases in victory probability are found. The strong advantages provided by successful use of resource towards the `endgame', where a successful gamble will produce instant victory or avoid instant defeat, shapes the structure of the optimal policy landscape. When little resource is available, complex structures emerge in the optimal landscape that depend on the tradeoff between using resource in the current state or `saving' it in case of a more beneficial state later (`risking points is not the same as risking the probability of winning' \cite{neller2004optimal}). When default victory is unlikely, using resource to reinforce rare success probabilities is a favoured strategy; when default victory is likely, using resource to mitigate rare loss probabilities is favoured. The specific optimal policy in a given state is solved and can be reasonably approximated by more heuristic strategies \cite{smith2007dynamic}. Interestingly, there is little performance loss when these heuristics are `open-loop', in the sense that policy choice only depends on a round's outcome and coarsely on the amount of current resource. `Feedback' strategies additionally based on the statistics of the two players did not provide a substantial advantage as long as endgame dynamics were covered.

Numerous FF gamebooks embellish the basic combat system, where weapons, armour, and other circumstances led to different \igjstamina{} costs or different rules. In the notoriously hard `Crypt of the Sorcerer' \cite{cryptbook}, if the final opponent (with $k_o = 12$) wins two successive rounds the player is instantly killed, altering the Markovian nature of the combat system (and substantially decreasing victory probability). A simulation study \cite{crypt}, while not employing an optimised \igjluck{} strategy, estimated a 95\% probability of this instant death occurring (and a 0.11\% probability of overall victory through the book). Further expansion of this analysis will generalise the potential rulesets, and hence allow the identification of optimal strategies in more situations.

Another route for expansion involves optimising victory probability while preserving some statistics, for example enforcing that $s_h > s^*$ or $l > l^*$ at victory, so that some resource is retained for the rest of the adventure after this combat. Such constraints could readily be incorporated through an initial reallocating $v_p$ over system states in the dynamic programming approach (Fig. \ref{dyncartoon}(i)), or by expanding the definition of the score being optimised to include some measure of desired retention in addition to $v_p$.


In an era of artificial intelligence approaches providing effective but essentially uninterpretable strategies for complex games \cite{campbell2002deep, lee2016human, gibney2016google}, more targetted analyses still have the potential to inform more deeply about the mechanisms involved in these strategies. Further, mechanistic understanding makes successful strategies readily available and simple to implement in the absence of computational resource. We hope that this increased interpretability and accessibility both contribute to the demonstration of the general power of these approaches, and help improve the experience of some of the millions of FF players worldwide. 

\section*{Acknowledgments}

The author is grateful to Steve Jackson, Ian Livingstone, and the many other FF authors for creating this system and the associated worlds. The author also thanks Daniel Gibbs and Ellen R\o yrvik for productive discussions.  



\bibliographystyle{apalike}
\bibliography{ff}

\clearpage

\section*{Appendix}

\subsection*{Detailed structure of optimal policy space}

\textbf{Lower $\Delta_k$.} In this regime, for $s_h = 2$, there is a band of strategy 2, reflecting the fact that \igjluck{} is best spent attempting to avoid the fatality of a lost attack round. For certain $s_{o}$ regions, strategy 3 is optimal. For example, when $l$ is low, strategy 3 appears when $s_o = 3$ or $4$. Here, the probability of winning an attack round is very low, so even a small chance of a successful \igjluck{} test amplifying the outcome to a killing blow is worth taking. These bands of strategy 3 propagate as $l$ increases, occupying strips between $s_o = 3$ and $s_o = 4n$. One $s_{opp}$ value is omitted from each band, at $s_o = 2 + 4n$. It will be noticed that these values are those where, if \igjluck{} is not employed, a successful round ($s_o - 2$) will decrease $\lceil s_o / 4 \rceil$, whereas if \igjluck{} is employed and is not successful ($s_o - 1$), that next band will not be reached. When \igjluck{} is a limited resource it does not pay to `spend' it in these cases.

Where strategy 2 and 3 bands overlap, intuitively, strategy 1 appears, using \igjluck{} for either outcome. As $l$ increases further ($l \geq 8$) other regions of \igjstamina{} space display a preference for strategy 2, reflecting the fact that substantial \igjluck{} is now available relative to the likely number of remaining attack rounds, so that is pays to invest \igjluck{} defensively as well as offensively. As $l$ increases further still, the expanding regions of strategy 2 overlap more with existing strategy 3 regions, leading to an expansion of strategy 1 in low $s_h$ regions.  

Column-wise, beginning at $\Delta_k = -6$, some horizontal bands in the the strategy 3 region sparsify into $s_h \Mod{2} = 1$ bands. This banding reflects the ubiquitous presence of strategy 2 when $s_h = 2$, to potentially defer the final fatal loss and allow one more attack round. If $s_h \Mod{2} = 1$, this region will not be reached unless the player uses a (non-optimal) defensive strategy elsewhere, so \igjluck{} can be freely invested in offensive strategy. If $s_h \Mod{2} = 0$, the battle has some probability (high if $\Delta_k$ is low) of encountering the $s_h = 2$ state, so there is an advantage to preserving \igjluck{} for this circumstance.

\textbf{Higher $\Delta_k$.} Patterns of strategies 2-3 remains largely unchanged as $\Delta_k$ increases, until the regions of strategy 3 at lower $l$ start to become sparsified, breaking up both row-wise and column-wise. Row-wise breaks, as above, occur at $s_o = 2+4n$, so that low \igjluck{} is not used if the outcome will not cross a $\lceil s_o / 4 \rceil$ band. Eventually these breaks are joined by others, so that for e.g. $\Delta_k = -4$, $l = 4$ strategy 3 only appears in bands of $s_o = 3 + 4n$. Here, a successful \igjluck{} test shifts the system to the next $\lceil s_o / 4 \rceil$ band, while an unsuccessful \igjluck{} test means that the next band can be reached with a successful attack round without using \igjluck{}. Column-wise, strategy 2 expands in $s_h \Mod{2} = 0$ bands at higher $\Delta_k$, and the ongoing sparsification of strategy 3 regions for low $l$ as $\Delta_k$ increases.

For higher $l$, as $\Delta_k$ increases, the sparsification of strategy 3 regions occurs in tandem with an expansion of strategy 2 regions. Strategy 2 becomes particularly dominant in high $s_o$, low $s_h$ regions. Here, the player expects to win most attack rounds, and the emphasis shifts to minimising losses from the rare attack rounds that the opponent wins, so that the player has time to win without their \igjstamina{} running out. A banded structure emerges (for example, $\Delta_k = 2, l=12$) where for $s_o \Mod{4} \leq 1$ strategy 2 is favoured and for $s_o \Mod{4} \geq 2$ strategies 1 and 3 are favoured. This structure emerges because of the relatively strong advantage of a successful \igjluck{} test at $s_o = 3$ and $s_o = 4$, leading to immediate victory, and the propagating advantage of successful \igjluck{} tests that lead to this region (wherever $s_o \Mod{4} \geq 2$). In the absence of this advantage ($s_o \Mod{4} \leq q$), the best strategy is to minimise damage from rare lost rounds until the advantageous bands are obtained by victories without depleting \igjluck{}.

At higher yet $\Delta_k$, victory is very likely, and the majority of \igjstamina{} space for high $l$ is dominated by the defensive strategy 2 -- minimising the impact of the rare attack rounds where the opponent wins. At intermediate $l$, strategies 2 and 3 appear in banded formation, where defensive \igjluck{} use is favoured when $s_h$ is even, to increase the number of steps needed to reach $s_h = 0$.

\renewcommand{\figurename}{Supplementary Figure}

\end{document}